\documentclass{llncs}

\usepackage{eccv}

\usepackage{eccvabbrv}

\usepackage{graphicx}
\usepackage{booktabs}
\graphicspath{{images/}}
\usepackage[accsupp]{axessibility}  

\usepackage{hyperref}
\usepackage{orcidlink}
\usepackage{multirow}
\usepackage{graphicx}
\usepackage[dvipsnames, table]{xcolor} 
\usepackage{amssymb} 
\usepackage{amsmath} 
\newcommand{\up}[1]{\textsubscript{\textcolor{red}{$\uparrow$#1}}}
\newcommand{\down}[1]{\textsubscript{\textcolor{ForestGreen}{$\downarrow$#1}}}
\newcommand{\res}[2]{#1\up{#2}}

\begin{document}
\pagestyle{plain} 
\title{Delineating Knowledge Boundaries for Honest Large Vision-Language Models
} 

\titlerunning{Abbreviated paper title}

\author{
    Junru Song \and 
    Yimeng Hu \and 
    Yijing Chen \and 
    Huining Li \and 
    Qian Li\thanks{Corresponding authors.} \and  
    Lizhen Cui\protect\footnotemark[1] \and     
    Yuntao Du\protect\footnotemark[1]          
}
\institute{Shandong University}

\maketitle

\begin{abstract}
Large Vision-Language Models (VLMs) have achieved remarkable multimodal performance yet remain prone to factual hallucinations, particularly in long-tail or specialized domains. Moreover, current models exhibit a weak capacity to refuse queries that exceed their parametric knowledge. In this paper, we propose a systematic framework to enhance the refusal capability of VLMs when facing such unknown questions. We first curate a model-specific "Visual-Idk" (Visual-I don't know) dataset, leveraging multi-sample consistency probing to distinguish between known and unknown facts. We then align the model using supervised fine-tuning followed by preference-aware optimization (e.g., DPO, ORPO) to effectively delineate its knowledge boundaries. Results on the Visual-Idk dataset show our method improves the Truthful Rate from 57.9\% to 67.3\%. Additionally, internal probing also demonstrates that the model genuinely recognizes its boundaries instead of just memorizing refusal patterns. Our framework further generalizes to out-of-distribution medical and perceptual domains, providing a robust path toward more trustworthy and prudent visual assistants.
\end{abstract}

\section{Introduction}
\label{sec:intro}

\begin{figure*}[t]
    \centering
    \includegraphics[width=1.0\linewidth]{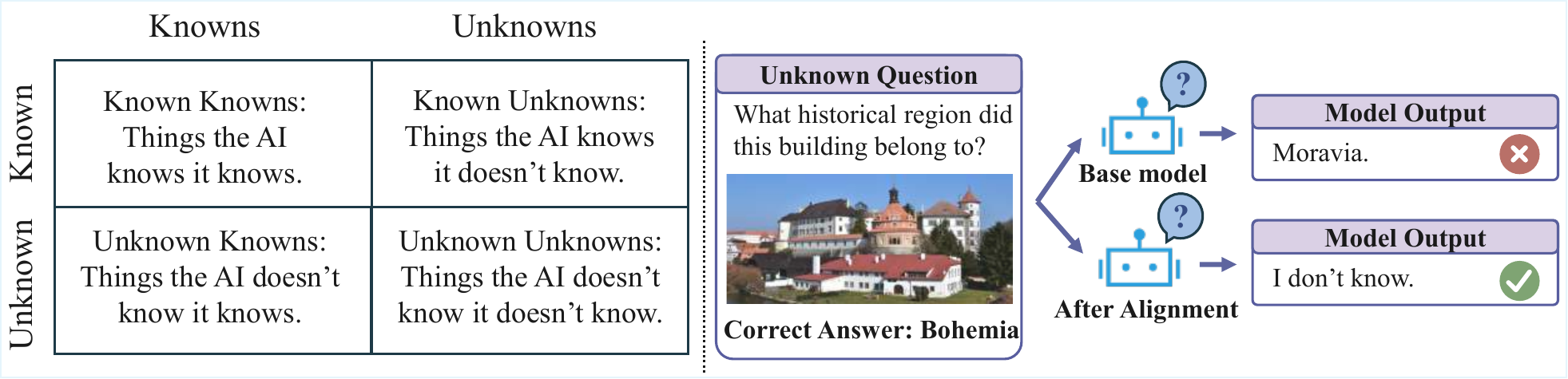} 
    \caption{\textbf{Left:} The four knowledge quadrants of a VLM, adapted from~\cite{cheng24Idk}. Our goal is to transform Unknown Unknowns (hallucinations) into Known Unknowns (standard refusals). \textbf{Right:} Comparison of model behaviors on an unknown question. The model after alignment correctly identifies its internal knowledge gap and provides a truthful refusal (``I don't know''), while the base model generates an incorrect response (``Moravia'').}
    \label{fig:overview}
    \vspace{-3mm}
\end{figure*}

The emergence of Large Vision-Language Models (VLMs) has marked a transformative milestone in multimodal artificial intelligence, enabling AI assistants to engage in complex reasoning and open-ended dialogue based on visual inputs\cite{Liu2023VisualIT}. Despite their impressive performance, these models are still highly prone to factual hallucinations\cite{Li2023EvaluatingOH}. In many cases, when confronted with questions involving images that contain long-tail entities or specialized domains beyond the model’s parametric knowledge, such as rare biological species or complex medical radiographs, VLMs tend to produce confident yet entirely incorrect responses rather than acknowledging their knowledge limitations~\cite{Chen2023CanPV}.
%
%
Such ``blind confidence'' poses significant risks in high-stakes applications like medical diagnosis or legal document analysis, where the cost of misinformation is prohibitively high\cite{Huang2023ASO}. Consequently, a fundamental question arises for the more responsible AI: \textbf{Can visual assistants discern their own knowledge boundaries and articulate this awareness through natural language?}

Existing research on mitigating VLM hallucinations has primarily focused on reducing object-level misidentifications\cite{xie2024v} or improving cross-modal alignment through reinforcement learning from human feedback (RLHF)\cite{yu2024rlhf}. While some studies\cite{cheng24Idk} in the LLM domain have explored the ``I don't know'' (Idk) response for pure text, the multimodal landscape remains largely unaddressed. Current VLM ``refusal'' studies often focus on perceptual uncertainty, where images are intentionally blurred or corrupted\cite{Gurari2018VizWizGC}. However, we argue that the more insidious challenge lies in \textbf{epistemic uncertainty}, where the visual input is perfectly clear, but the model lacks the internal parametric knowledge to interpret it\cite{wang2024drawing}. As illustrated in the knowledge quadrants of \textbf{Fig.~\ref{fig:overview} (left)}, vanilla VLMs often fall into the \textit{Unknown Unknowns} (IDK-IDK) segment, where they generate confident but fabricated claims for entities they do not actually master. Current alignment paradigms emphasize ``helpfulness'' at the expense of ``truthfulness''\cite{Lin2021TruthfulQAMH}, inadvertently forcing models to guess in the absence of evidence.

To overcome this challenge, In this paper, we propose the \textbf{Visual-Idk alignment framework}, a systematic pipeline to calibrate Visual Assistants with their internal knowledge boundaries. Our methodology consists of three core stages: (i) Visual-Semantic Sample Selection, starting with an existing dataset, we utilize a powerful VLM to filter low-quality samples; (ii) Visual Knowledge Probing via multi-sample consistency to quantify parametric mastery; and (iii) Preference Pair Generation, which constructs contrasting samples to supervise the model's decision-making. We then employ preference-aware optimization (e.g., DPO, ORPO) to reshape the model's epistemic boundaries, cultivating a more truthful and prudent assistant.

We conduct extensive experiments, and the results on the V-Idk test set show that the framework significantly enhances reliability, improving the \textsc{Truthful} rate of LLaVA-1.5-7B from \textbf{57.9\% to 67.3\%}. Notably, ORPO achieves a superior balance and preserves mastered knowledge (IK-IK) more effectively than Supervised Fine-Tuning (SFT). Besides, the proposed framework is domain-agnostic. It could generalize robustly to ScienceQA~\cite{lu2022learn} dataset as well as OOD scenarios such as blurred images (VizWiz \cite{Gurari2018VizWizGC}) and specialized medical imagery (PMC-VQA\cite{zhang2023pmcvqa}). The generalization results suggest that the alignment internalizes a fundamental awareness of knowledge mastery rather than overfitting to specific entities. Furthermore, internal probing of model Logprob\cite{hills2023usinglogprobs} and PPL confirms that the model truly learns to recognize its own limits instead of just memorizing a response pattern.

To sum up, our contributions are summarized as follows:

    \textbf{(1)} We introduce a framework that effectively improves epistemic uncertainty by teaching models to express unknowns for the question that goes beyond parametric knowledge.
    
    \textbf{(2)} We propose a novel pipeline for constructing knowledge-intensive preference datasets, specifically designed to identify the epistemic boundaries of VLMs regarding their parametric knowledge.
    
    \textbf{(3)} We conduct extensive experiments on multiple datasets, and the results show that the proposed method could improve truthfulness effectively.


\vspace{-3mm}
\section{Related Work}
\label{sec:related_work}

\vspace{-3mm}
\subsection{Knowledge Boundary}
Despite the rapid advancement of Large Vision-Language Models (VLMs) such as LLaVA\cite{liu2023visual}, InstructBLIP\cite{dai2023instructblip}, and Qwen-VL\cite{wang2024qwen2}, factual hallucination remains a persistent bottleneck. Prior work commonly categorizes VLM hallucinations into object-level errors (e.g., mentioning nonexistent objects) and attribute-level errors (e.g., incorrect colors, counts, or spatial relations). To mitigate these issues, existing approaches have explored (i) scaling up high-quality multimodal instruction data\cite{liu2023mitigating}, (ii) post-hoc verification with external tools or retrieval modules\cite{yin2024woodpecker}, and (iii) architectural or training refinements that improve cross-modal alignment\cite{yu2024rlhf}. While effective to varying degrees, these methods largely share a common goal: making models answer more accurately. Yet, an equally important capability is often overlooked—knowing when not to answer. In practice, VLMs are frequently incentivized to guess plausible responses even when confronted with long-tail or specialized entities that lie beyond their parametric knowledge, resulting in confident but unfounded claims. This motivates a complementary perspective: aligning VLMs to be explicitly aware of their knowledge boundaries and to appropriately abstain when visual evidence is clear but the required world knowledge is missing or uncertain.

The pursuit of honest AI has been more systematically investigated in the text-only domain, where researchers have shown that LLMs can exhibit an intrinsic ability to assess the likelihood of their own claims. Building on this observation, Cheng et al.\cite{cheng24Idk} proposed the “Idk” (I don’t know) framework, which teaches LLMs to recognize and verbalize their knowledge limits by constructing model-specific datasets and applying supervised fine-tuning (SFT) followed by preference optimization. Related efforts, including R-Tuning\cite{zhang2024r} and Hindsight Instruction Relabeling\cite{zhang2023wisdom}, further explored learning-to-refuse behaviors for unanswerable or underspecified queries. Our work follows this technical lineage but extends it to the multimodal setting, where refusal decisions depend on a more entangled interplay between visual perception (what is present in the image) and internal knowledge mastery (what the model actually knows about what it sees). Different from InBoL \cite{wang2024drawing}, which explores generic knowledge boundaries by focusing on questions that are unanswerable by nature, our framework targets epistemic gaps in more knowledge-intensive scenarios involving long-tail and professional domains. Consequently, we focus on cultivating VLMs’ self-awareness of knowledge boundaries—not merely improving answer quality, but enabling prudent abstention when the query exceeds the model’s reliable knowledge.

\subsection{Preference-Aware Optimization}


Direct Preference Optimization (DPO)~\cite{rafailov2023direct} and its variants (e.g., ORPO~\cite{hong2024orpo}, CPO~\cite{tan2023co}) have emerged as practical alternatives to reinforcement learning from human feedback (RLHF) by directly optimizing models from pairwise preference data without the need for explicit reward modeling. Compared with traditional RLHF pipelines that require reward model training and reinforcement learning stages, preference-based optimization provides a more stable and computationally efficient framework for aligning large models.

In the text-only domain, the preference optimization has been widely adopted to align helpfulness and reduce unsafe or untruthful model behaviors \cite{cheng24Idk}\cite{brahman2024art}. In the multimodal setting, recent studies have also explored preference-based alignment for VLMs, improving instruction-following and factuality and thereby reducing hallucinations\cite{xie2024v}\cite{ouali2024clip}. Unlike existing VLM preference-optimization that focuses on answer quality, we target the answer–refusal boundary to mitigate over-refusal. We build upon these preference-aware techniques to explicitly calibrate the answer–refusal boundary, encouraging truthful abstention under epistemic uncertainty while preserving helpfulness on answerable queries, thereby alleviating the over-refusal (alignment tax) often observed in SFT-only refusal tuning.


\section{The Visual-Idk Alignment Framework}

\label{sec:method}

In this section, we detail our framework for aligning Visual Assistants to recognize their knowledge boundaries. Our approach follows a transition from a non-training baseline to sophisticated training-based alignment, aimed at calibrating the model's epistemic uncertainty.

\subsection{Construction of the Visual-Idk Dataset}
\label{subsec:dataset}
\begin{figure}[t]
    \centering
    \includegraphics[width=\linewidth]{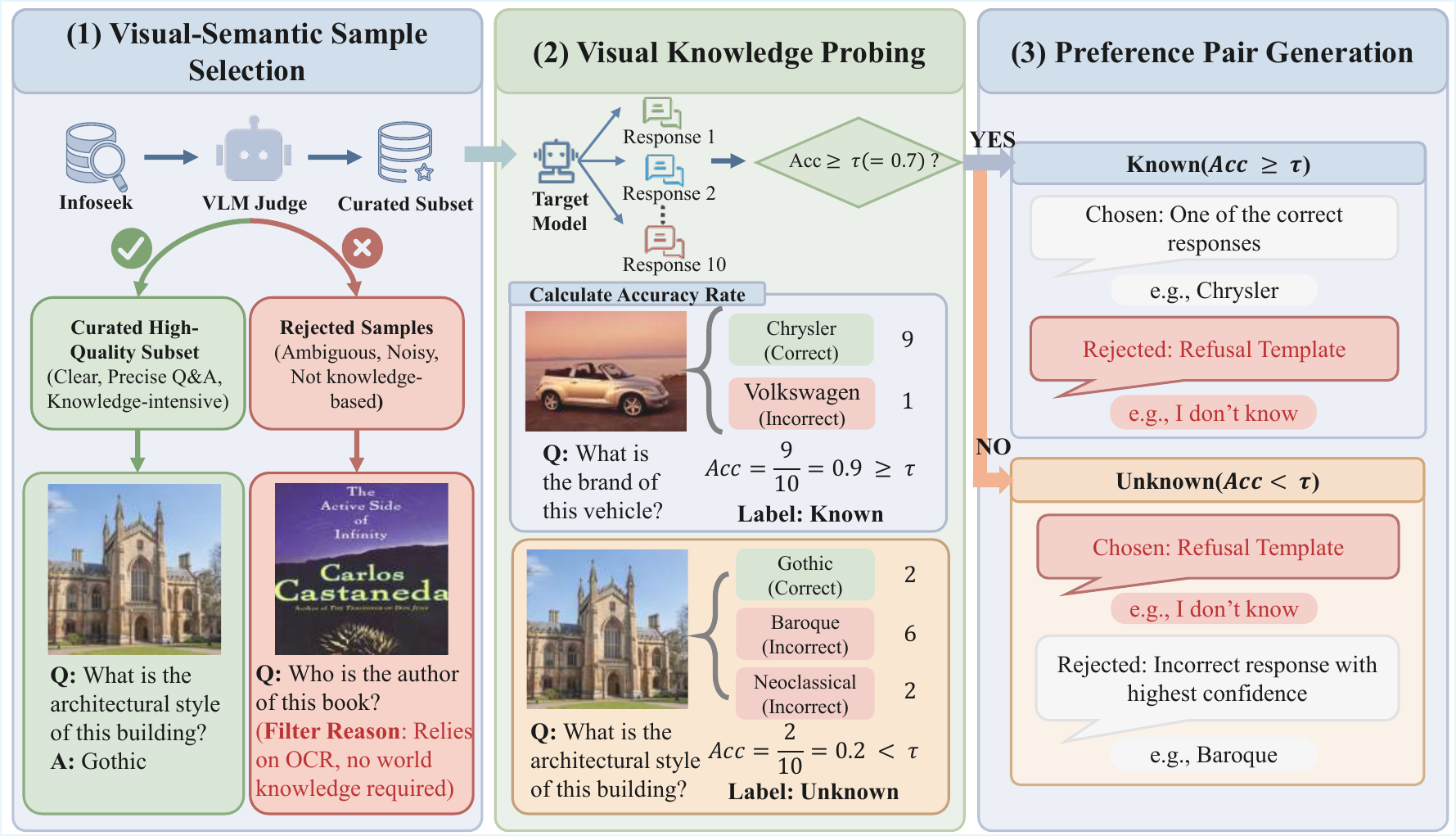} 
    \caption{\textbf{The pipeline of Visual-Idk dataset construction.} The process comprises three major stages: (1) Visual-Semantic Sample Selection for quality control; (2) Visual Knowledge Probing to identify model-specific knowledge boundaries; and (3) Preference Pair Generation.}
    \label{fig:pipeline}
\end{figure}

The foundation of our approach is the creation of a model-specific Visual-Idk dataset. As illustrated in Fig. \ref{fig:pipeline}, this process identifies the specific parametric knowledge gaps of the target model while ensuring data quality.

\textbf{Visual-Semantic Sample Selection.} To isolate \textit{epistemic uncertainty} from \textit{perceptual errors}, we first perform a high-quality sample selection process from an existing dataset (Stage 1 in Fig. 2). We leverage a high-capability VLM model (e.g., Qwen2.5-VL) as a semantic judge to curate image-question pairs from the initial InfoSeek dataset. We retain only those samples meeting three specific criteria: (1) visual quality is clear and unambiguous, (2) the ground-truth answer is precise, and (3) the query is knowledge-intensive rather than relying on basic optical character recognition (OCR) or simple perception. This rigorous selection ensures that model failures on the resulting curated subset can be reliably attributed to parametric knowledge gaps.

\textbf{Visual Knowledge Probing.} On the filtered dataset, we assess the target model $\pi_\theta$ via an empirical probing process (Stage 2 in Fig. \ref{fig:pipeline}). We sample $N=10$ independent responses for each image-question pair $(x, q)$ using a high decoding temperature. We define a mastery threshold $\tau$ (the \textit{Ik threshold}). A sample is categorized as known if its average accuracy rate of 10 responses $A(x, q) \ge \tau$, indicating the model can consistently retrieve the correct fact. Otherwise, it is labeled as Unknown.

\textbf{Preference Pair Generation.} Based on the probing results, we construct the V-Idk training set $\mathcal{D}_{V-Idk}$ by generating preference pairs $(y_w, y_l)$ to contrast desirable behaviors (Stage 3 in Fig. \ref{fig:pipeline}):

    \textbf{ (1) For Known Facts ($A \ge \tau$):} To prevent over-conservatism (the ``alignment tax''), we set the correct response as the chosen $y_w$ and a refusal template (e.g., ``I don't know'') as the rejected $y_l$.
    
   \textbf{ (2) For Unknown Facts ($A < \tau$):} To suppress hallucinations, we set the refusal template as the chosen $y_w$ and the model's previously generated incorrect response with the highest confidence as the rejected $y_l$.

\textbf{Dataset Statistics and Source.} 
We construct the Visual-Idk dataset based on the InfoSeek benchmark \cite{chen2023infoseek}. We prioritize InfoSeek over traditional VQA datasets (e.g., VQA v2 \cite{goyal2017making} or OK-VQA \cite{marino2019ok}) due to its extreme knowledge-intensity and long-tail entity distribution. While the original InfoSeek benchmark contains a certain degree of perceptual noise and ambiguous queries, our visual-semantic sample selection ensures that the resulting curated subset maintains high visual and textual quality. This refinement renders the data highly conducive to the task of knowledge boundary identification: by minimizing visual ambiguity and reasoning noise, model failures on this subset can be more reliably attributed to gaps in internal parametric knowledge rather than external perceptual errors. 

The finalized V-Idk dataset comprises 20,000 training samples and 3,000 test samples. Regarding its linguistic characteristics, the questions are designed to be concise and information-dense, with a mean length of approximately 10 words, while the corresponding ground-truth answers are highly succinct, averaging 3 to 4 words. Based on our multi-sample consistency probing (with $\tau=0.7$), the dataset is curated to maintain a specific distribution: approximately 60\% of the samples are categorized as ``Known'' (IK), while the remaining 40\% are ``Unknown'' (IDK). This balanced ratio ensures that the alignment process provides sufficient supervision for both preserving helpfulness on mastered facts and suppressing hallucinations on unknown entities.

\subsection{Experimental Methods}

We include several popular methods. The first is simple prompting. Besides, we also investigate four training strategies to explicitly align the model with its knowledge boundaries using the V-Idk dataset.

\textbf{Idk-Prompting}. As a zero-training baseline, we evaluate the model's latent self-awareness via Idk-Prompting. We prepend an explicit instruction to the input: \textit{``Answer the following question based on the image. If you do not know the answer, please respond with `I'm sorry, this question is beyond my knowledge. I don't know the answer.' ''} This strategy relies purely on the model's pre-trained instruction-following capability without any parameter updates.


\textbf{Supervised Fine-tuning (SFT).} We first perform Idk-SFT by maximizing the log-likelihood of the chosen responses $y_w$ in $\mathcal{D}_{V-Idk}$:
\begin{equation}
\mathcal{L}_{SFT} = -\mathbb{E}_{(x, q, y_w) \sim \mathcal{D}_{V-Idk}} \left[ \log \pi_\theta(y_w | x, q) \right]
\end{equation}
SFT serves as the first stage of training to teach the model the standardized refusal format.

\textbf{Direct Preference Optimization (DPO).} To further calibrate the decision boundary, we use DPO to contrast honest refusals with hallucinations. The loss is defined as:
\begin{equation}
\mathcal{L}_{DPO} = -\mathbb{E} \left[ \log \sigma \left( \beta \log \frac{\pi_\theta(y_w|x, q)}{\pi_{ref}(y_w|x, q)} - \beta \log \frac{\pi_\theta(y_l|x, q)}{\pi_{ref}(y_l|x, q)} \right) \right]
\end{equation}
where $\pi_{ref}$ is the model after SFT. 

\textbf{Contrastive Preference Optimization (CPO).} We also evaluate CPO, which avoids the need for a reference model by directly optimizing a contrastive objective between $y_w$ and $y_l$. It encourages the model to stay close to the chosen distribution while moving away from the rejected one.

\textbf{Odds Ratio Preference Optimization (ORPO).} Finally, we utilize ORPO to combine supervised learning and preference alignment into a single objective:
\begin{equation}
\mathcal{L}_{ORPO} = \mathcal{L}_{SFT} + \lambda \cdot \mathcal{L}_{OR}, \quad \mathcal{L}_{OR} = -\log \sigma \left( \log \frac{odd_{\pi_\theta}(y_w|x, q)}{odd_{\pi_\theta}(y_l|x, q)} \right)
\end{equation}
where $odd_\pi(y) = \frac{\pi(y)}{1-\pi(y)}$. ORPO regularizes the model to increase the relative odds of being honest for unknown facts while remaining helpful for known facts.

\subsection{Evaluation Metrics}
\label{subsec:metrics}

To formalize the assessment of a model's knowledge boundaries, we categorize the aligned model's inference behaviors into four quadrants, as illustrated in \textbf{Fig.~\ref{fig:overview} (left)}. These quadrants, \textit{Known Knowns}, \textit{Known Unknowns}, \textit{Unknown Knowns}, and \textit{Unknown Unknowns}, provide a granular mapping of the model's epistemic state by intersecting its initial mastery level $A(x, q)$ (from Sec. 3.1) with the semantic correctness of its output during testing. We define the evaluation dataset $\mathcal{D}$ and denote the subset of initially unmastered samples (the ``Unknowns'' column in \textbf{Fig.~\ref{fig:overview}}) as the \textit{Unknown} subset $\mathcal{D}_{unk} = \{(x, q) \in \mathcal{D} \mid A(x, q) < \tau\}$.

Given that model responses are in natural language, we utilize a high-capability LLM as a semantic judge to evaluate model behavior. We define a logical verification predicate $\mathcal{V}(y_{pred}, \text{target})$ that returns \textit{True} if the semantic content of $y_{pred}$ aligns with the intended \textit{target} (either the ground-truth answer $y_{gt}$ or a refusal intent). Based on this, we define three key metrics:

\textbf{IK-IK Rate ($\rho_{\text{IK}}$) (Knowledge Accuracy)}: Quantifies the model's effective factual knowledge base during inference. As visually represented by the transition in \textbf{Fig.~\ref{fig:overview} (right)}, a sample is classified as IK-IK if the model provides a correct factual response, regardless of whether the sample was initially categorized as Known or Unknown. It is defined as:
    \begin{equation}
        \rho_{\text{IK}} = \frac{\sum_{(x,q) \in \mathcal{D}} \mathbb{I}\left( \mathcal{V}(y_{pred}, y_{gt}) = \text{True} \right)}{|\mathcal{D}|}
    \end{equation}
    where $\mathbb{I}(\cdot)$ is the indicator function. This metric reflects the model's ability to retrieve and articulate correct facts across the entire domain.

 \textbf{IK-IDK Rate ($\rho_{\text{IDK}}$) (Admitting Ignorance)}: Measures the model's self-awareness by quantifying its ability to correctly refuse queries for which it lacks sufficient parametric knowledge ($A(x, q) < \tau$). Conceptually, this represents the model successfully landing in the \textit{Known Unknown} quadrant (the top-right cell of \textbf{Fig.~\ref{fig:overview}}). It is defined as:
    \begin{equation}
        \rho_{\text{IDK}} = \frac{\sum_{(x,q) \in \mathcal{D}_{unk}} \mathbb{I}\left( \mathcal{V}(y_{pred}, \text{refusal}) = \text{True} \right)}{|\mathcal{D}|}
    \end{equation}

 \textbf{\textsc{Truthful} Rate ($\mathcal{T}$) (Overall Reliability)}: Our primary evaluation metric, representing the total frequency of ``honest'' behaviors, either providing a correct factual answer or correctly admitting ignorance when mastery is low:
    \begin{equation}
        \mathcal{T} = \rho_{\text{IK}} + \rho_{\text{IDK}}
    \end{equation}
    Maximizing the \textsc{Truthful} rate is the central objective of our alignment framework, as it captures the optimal balance between being helpful and being honest.

Consistent with this framework, the remaining behaviors are classified as IDK-IK (the ``alignment tax,'' where the model refuses a known fact) and IDK-IDK (hallucinations, where the model provides an incorrect answer to an unknown fact, illustrated by the Base Model in \textbf{Fig.~\ref{fig:overview} (right)}). All metrics are normalized by the total dataset size $|\mathcal{D}|$.

\section{Experiments}

In this section, we evaluate the efficacy of our framework in aligning Visual Assistants with their internal knowledge boundaries. We present the results across five experimental dimensions: (1) main performance on V-Idk, (2) cross-dataset generalization on ScienceQA, (3) Generalization to out-of-distribution (OOD) scenarios, (4) internal probabilistic probing, and (5) case study.

\subsection{Implementation Details}
\label{subsec:implementation}

\textbf{Training Configurations.} 
We implement our framework using PyTorch and the DeepSpeed library. All models are trained on a server equipped with 2 NVIDIA H800 (80G) GPUs. We adopt Low-Rank Adaptation (LoRA) \cite{hu2021lora} for parameter-efficient fine-tuning, targeting all linear modules in the LLM backbone with a rank $r=64$ and $\alpha=128$.

The alignment process consists of two stages:
(1) Supervised Fine-tuning (SFT): The model is trained on the curated 20,000 V-Idk training samples for 1 epoch. We employ the AdamW optimizer with a learning rate of $1 \times 10^{-5}$ and a global batch size of 32. (2) Preference Optimization: For DPO, CPO, and ORPO stages, the model is further aligned for 3 epochs using 20,000 preference pairs. The learning rate is reduced to $1 \times 10^{-6}$ with a cosine decay scheduler, maintaining the same LoRA configuration ($r=64$).

\textbf{Inference and Evaluation.} 
During inference, we evaluate the aligned models on the 3,000 V-Idk test samples with a distribution of 60\% Known (IK) and 40\% Unknown (IDK) samples. We utilize greedy decoding to ensure deterministic results. As established in Sec. \ref{subsec:dataset}, each test sample is categorized into knowledge quadrants by comparing the model's output against the ground-truth and refusal templates via a high-capability LLM judge. To assess robustness, we report results under both \textit{zero-shot} and \textit{few-shot} settings as described in Sec. \ref{subsec:main_results}.

\subsection{Main Results on V-Idk}
\label{subsec:main_results}

\begin{table*}[t]
\centering
\caption{\textbf{Main Results on V-Idk.} Comparison of alignment methods on LLaVA-1.5 models across two inference settings. Subscripts denote improvement (\textcolor{red}{$\uparrow$}) or decrement (\textcolor{ForestGreen}{$\downarrow$}) over the baseline.}
\label{tab:main_results_enhanced}
\resizebox{\textwidth}{!}{
\setlength{\tabcolsep}{5pt}
\begin{tabular}{l|l|lll|lll}
\toprule
\multirow{2.5}{*}{\textbf{Model}} & \multirow{2.5}{*}{\textbf{Method}} & \multicolumn{3}{c|}{\textbf{Zero-shot}} & \multicolumn{3}{c}{\textbf{Few-shot}} \\
\cmidrule(lr){3-5} \cmidrule(lr){6-8}
&  & IK-IK & IK-IDK & \textsc{Truthful} & IK-IK & IK-IDK & \textsc{Truthful} \\
\midrule
\multicolumn{2}{l|}{\textit{\textbf{Data Proportion}}} & \textit{60.0} & \textit{40.0} & \textit{100.0} & \textit{60.0} & \textit{40.0} & \textit{100.0} \\
\midrule
\multirow{5}{*}{\textbf{LLaVA-1.5 7B}}
& Base & 45.8 & 1.8  & 47.6 & \textbf{45.5} & 12.4 & 57.9 \\
& SFT       & 34.4 \down{11.4} & \textbf{32.7} \up{30.9} & \textbf{67.1} \up{19.5} & 27.6 \down{17.9} & \textbf{36.8} \up{24.4} & 64.4 \up{6.5} \\
& DPO       & \textbf{51.9} \up{6.1} & 5.5 \up{3.7} & 57.4 \up{9.8} & 45.3 \down{0.2} & 17.8 \up{5.4} & 63.1 \up{5.2} \\
& CPO       & 44.4 \down{1.4} & 13.9 \up{12.1} & 58.3 \up{10.7} & 42.8 \down{2.7} & 24.5 \up{12.1} & \textbf{67.3} \up{9.4} \\
& ORPO      & 45.9 \up{0.1} & 14.1 \up{12.3} & 60.0 \up{12.4} & 40.0 \down{5.5} & 26.5 \up{14.1} & 66.5 \up{8.6} \\
\midrule·
\multirow{5}{*}{\textbf{LLaVA-1.5 13B}}
& Base & 49.9 & 4.3 & 54.2 & 49.6 & 19.8 & 69.4 \\
& SFT       & 34.4 \down{15.5} & \textbf{39.1} \up{34.8} & \textbf{73.5} \up{19.3} & 27.5 \down{22.1} & \textbf{39.8} \up{20.0} & 67.3 \down{2.1} \\
& DPO       & \textbf{50.2} \up{0.3} & 6.7 \up{2.4} & 56.9 \up{2.7} & \textbf{49.7} \up{0.1} & 21.4 \up{1.6} & 71.1 \up{1.7} \\
& CPO       & 44.9 \down{5.0} & 19.0 \up{14.7} & 63.9 \up{9.7} & 43.7 \down{5.9} & 30.3 \up{10.5} & 74.0 \up{4.6} \\
& ORPO      & 45.8 \down{4.1} & 20.4 \up{16.1} & 66.2 \up{12.0} & 42.7 \down{6.9} & 33.9 \up{14.1} & \textbf{76.6} \up{7.2} \\
\bottomrule
\end{tabular}
}
\end{table*}

We evaluate the performance of various alignment methods on the V-Idk test dataset under two distinct inference protocols to assess the model's robustness in discerning knowledge boundaries:
\textbf{(1) Zero-shot}: The model is provided only with the image and the query. No explicit instructions to abstain are given, nor are any refusal templates provided in the prompt. This setting evaluates the \textit{intrinsic} truthfulness and self-awareness internalized by the model during the alignment process.
    \textbf{(2) Few-shot}: The model is explicitly instructed: \textit{``If you do not know the answer, please respond with `I don't know'.''} Furthermore, two in-context demonstrations are provided: one showing a correct factual response to a known entity and another illustrating a standard refusal for an unknown entity. This setting measures the model's ability to perform \textit{guided} boundary calibration via in-context learning.

Table~\ref{tab:main_results_enhanced} compares the baseline \textit{Prompting} with four training-based strategies across LLaVA-1.5 7B and 13B models. Based on the results, we have the following observations:

\textbf{(1) Vanilla VLMs exhibit inherent overconfidence.} 
The \textit{Prompting} baseline in the zero-shot setting (7B) achieves a high IK-IK rate (45.8) but a negligible IK-IDK rate (1.8). This confirms that pre-trained VLMs, while knowledgeable, default to generating fabricated responses when encountering parametric gaps. Even under few-shot guidance, overconfidence remains a bottleneck, as the model continues to guess for the majority of unknown queries.

\textbf{(2) SFT enhances refusal capability but induces substantial suppression of mastered knowledge.} 
While SFT on V-Idk yields substantial gains in IK-IDK, it incurs a significant ``alignment tax.'' Specifically, for the 7B model, IK-IK accuracy decreases by $11.4\%$ in zero-shot and even more substantially by $17.9\%$ in the few-shot setting. This suggests that while SFT teaches a refusal format, it tends to over-generalize this behavior into a conservative prior, causing the model to erroneously reject queries it has actually mastered.

\textbf{(3) Preference optimization methods achieve a more calibrated knowledge–refusal balance.} 
Preference-aware methods (DPO, CPO, and ORPO) provide a superior trade-off compared to SFT. Instead of a blanket refusal prior, these methods learn to differentiate between mastery and ignorance. Among them, \textbf{ORPO} demonstrates the most prominent overall performance, achieving the leading \textsc{Truthful} scores in the 13B few-shot setting (76.6) and 7B zero-shot setting (60.0). This indicates that jointly optimizing for supervised likelihood and preference odds allows the model to curb hallucinations effectively while maintaining high factual utility.

\textbf{(4) Model scaling facilitates more precise boundary identification.} 
Larger models (13B) consistently outperform their 7B counterparts across both truthfulness and knowledge retention. This suggests that increased parametric capacity provides a more stable representation of internal knowledge, enabling the model to respond more robustly to alignment objectives.

\textbf{(5) Few-shot demonstrations facilitate superior truthfulness calibration.} 
Across all methods, Few-shot inference consistently outperforms Zero-shot in terms of \textsc{Truthful} rate. This improvement is attributed to the contextual anchors provided by the demonstrations, which help the model better calibrate its response threshold. Notably, the \textit{intrinsic honesty} of aligned models is evident: the ORPO-aligned 7B model in Zero-shot (60.0) even slightly exceeds the performance of the base model in the Few-shot setting (57.9), suggesting that the knowledge boundary has been successfully internalized.

\subsection{Insightful Analysis}

\label{subsec:scienceqa_evaluation}

\textbf{Cross-Dataset Evaluation on ScienceQA.}
To evaluate the cross-dataset generalization of the learned honest behavior, we directly evaluate the Visual-Idk-trained models on the ScienceQA benchmark \cite{lu2022learn}. This experiment assesses whether the alignment internalized from encyclopedic knowledge can successfully transfer to the distinct domain of scientific reasoning. 

The evaluation protocol remains strictly consistent with the main experiments: we first perform the same multi-sample consistency probe on the ScienceQA test set to identify the model's mastered and unmastered facts. This unified measurement allows for a direct assessment of how the model's awareness of its knowledge boundaries generalizes across different task distributions and data sources. Table~\ref{tab:scienceqa_truthfulness} summarizes the results for LLaVA-1.5 7B. We observe the following:

\textbf{(1) Alignment capability successfully transfers to scientific domains.} 
Despite being trained solely on V-Idk, all aligned methods demonstrate a superior ability to discern scientific knowledge boundaries compared to the baseline. ORPO achieves a leading \textsc{Truthful} score of 63.8\%, representing a 7.7\% improvement. This suggests that our framework calibrates a fundamental internal confidence that is not overfitted to specific entities but represents a domain-agnostic honesty that generalizes to complex scientific facts.

\textbf{(2) Consistent behavioral patterns in cross-dataset knowledge retention.} 
The behavioral trade-offs observed on ScienceQA echo our primary findings: SFT remains highly conservative, resulting in a high IK-IDK rate but the lowest IK-IK accuracy (24.0\%). Conversely, DPO stands out as the only method that effectively preserves and enhances scientific knowledge mastery, achieving the highest IK-IK rate of 33.9\%. ORPO and CPO strike the most effective balance, maintaining superior overall truthfulness while successfully curbing the ``alignment tax'' on previously unseen scientific knowledge.

\begin{table}[t]
\centering
\caption{\textbf{Cross-Dataset Evaluation on ScienceQA (Few-shot).} Comparison of alignment methods trained solely on Visual-Idk. The evaluation set is balanced with a 50\%:50\% ratio between Known and Unknown samples. Subscripts ($\uparrow$) denote the improvement over the baseline.}
\label{tab:scienceqa_truthfulness}
\setlength{\tabcolsep}{10pt}
\resizebox{0.9\columnwidth}{!}{
\begin{tabular}{l|lll}
\toprule
\textbf{Method} & \textbf{IK-IK} & \textbf{IK-IDK} & \textsc{\textbf{Truthful}} \\
\midrule
\textit{\textbf{Data Proportion}} & \textit{50.0\%} & \textit{50.0\%} & \textit{100.0\%} \\
\midrule
\textbf{Prompting} & 30.3 & 25.8 & 56.1 \\
\midrule
\textbf{SFT}   & {24.0}\down{6.3} & \res{35.5}{9.7} & \res{59.5}{3.4} \\
\textbf{DPO}   & \textbf{\res{33.9}{3.6}}   & \res{26.2}{0.4}   & \res{60.1}{4.0}  \\
\textbf{CPO}   & {26.8}\down{3.5}  & \textbf{\res{35.8}{10.0}} & \res{62.6}{6.5} \\
\textbf{ORPO}  & {28.1}\down{2.2} & \res{35.7}{9.9} & \textbf{\res{63.8}{7.7}} \\
\bottomrule
\end{tabular}
}
\vspace{-3mm}
\end{table}

\label{subsec:ood_generalization}

\textbf{Truthfulness Generalization on OOD Scenarios.}
To evaluate the generalization of the learned honesty, we extend our analysis to VizWiz-Unans \cite{Gurari2018VizWizGC} and PMC-VQA \cite{zhang2023pmcvqa}, representing visual and knowledge domain shifts, respectively. The VizWiz-Unans dataset consists of low-quality, blurry, or obscured real-world images that introduce significant perceptual uncertainty. Similarly, on PMC-VQA, the base model typically lacks the expert-level precision and factual grounding required for reliable responses. In these high-uncertainty scenarios—where forced responses inevitably lead to unfounded claims (hallucinations)—identifying mastery gaps and electing to abstain is the only truthful behavior. For these intrinsically unanswerable OOD queries, the \textsc{Truthful} rate is therefore defined strictly as the successful refusal rate.

\begin{figure}[t]
    \centering
    \includegraphics[width=\linewidth]{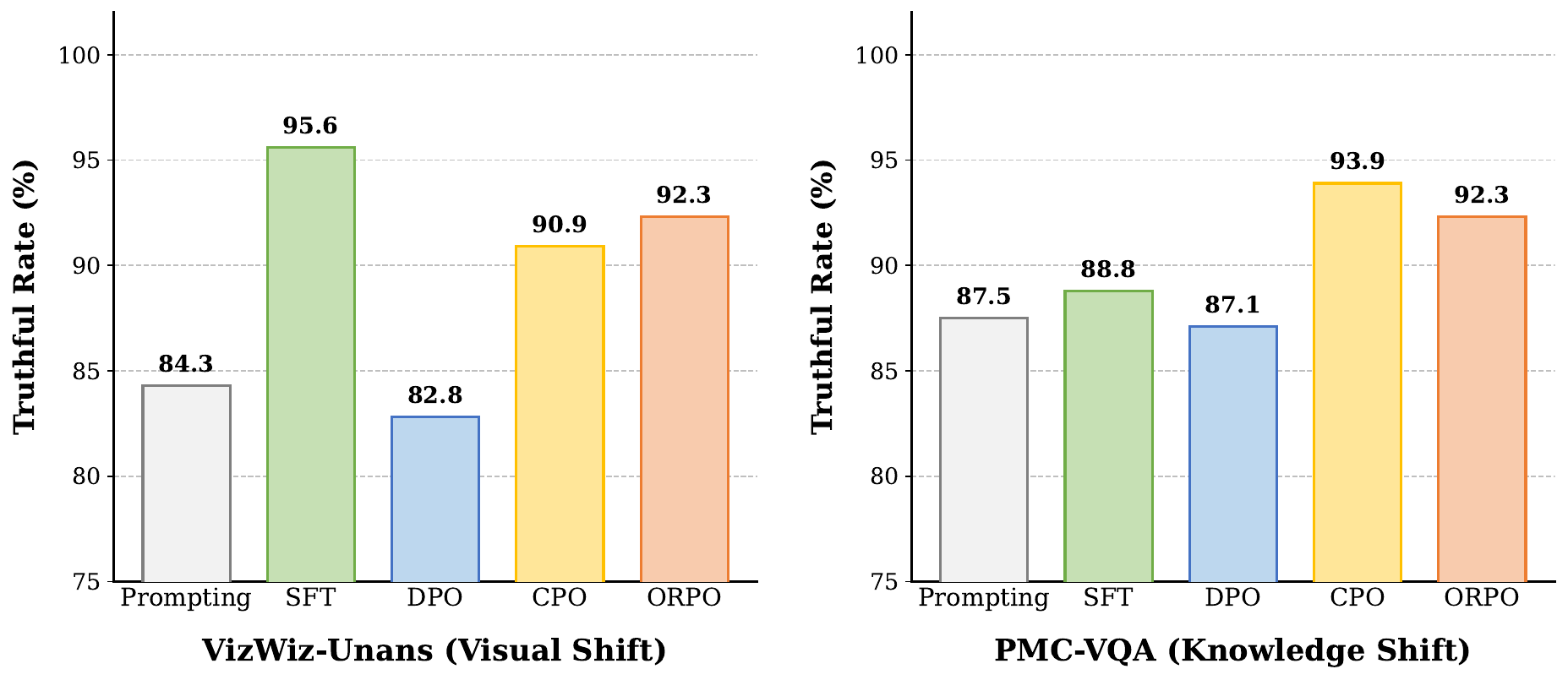}
    \caption{\textbf{Generalization on OOD scenarios.} Numeric labels denote the \textsc{Truthful} rate (\%).}
    \label{fig:ood_bar}
    \vspace{-3mm}
\end{figure}

As illustrated in Fig.~\ref{fig:ood_bar}, we observe the following:

\textbf{(1) SFT captures generic refusal patterns under visual noise.} 
On the VizWiz-Unans dataset, the SFT-aligned model achieves a peak \textsc{Truthful} rate of 95.6\%, a gain of 11.3\% over the baseline. This suggests that supervised fine-tuning is effective at establishing a conservative prior that prioritizes abstention when visual signals are severely degraded.

\textbf{(2) Preference-aware methods excel in specialized domain shifts.} 
While SFT performs well under visual noise, CPO and ORPO demonstrate superior calibration when encountering professional knowledge gaps. Specifically, CPO achieves the leading score of 93.9\% on PMC-VQA. This indicates that preference learning better enables the model to identify its lack of expertise in specialized fields like medicine, rather than merely relying on a fixed refusal template.

\textbf{(3) Performance degradation in DPO.} 
Notably, DPO exhibits a performance drop compared to the Prompting baseline in both OOD scenarios, falling to 82.8\% on VizWiz. This indicates that DPO may struggle to maintain robust refusal behavior when encountering extreme distribution shifts that were not adequately covered during preference training. In contrast, CPO and ORPO consistently maintain high truthfulness across both datasets.

\label{subsec:uncertainty_analysis}

\textbf{Internal Uncertainty Analysis.}
To delve into the model's internal state during the decision-making process, we perform a probabilistic probing analysis focused on the refusal response. Specifically, we force the model to generate the refusal template (``I don't know.'') across both ``Known'' (IK) and ``Unknown'' (IDK) subsets, then analyze the resulting average log-probabilities (\textbf{Logprob}\cite{hills2023usinglogprobs}) and \textbf{Perplexity (PPL)}. This allows us to discern whether the model's refusal is based on genuine epistemic calibration or mere template mimicry. Table~\ref{tab:uncertainty_analysis} summarizes the transition from \textit{Pre-Align} to \textit{Post-Align} behavior.

\textbf{(1) Internal Recalibration of Refusal Behavior.} 
After alignment, the model demonstrates a significant shift toward a more confident and stable refusal mode, especially for unmastered facts. For ``Unknown'' (IDK) questions, the model's confidence in admitting ignorance increases (Logprob: $-0.80 \to -0.63$), while its uncertainty regarding the refusal template reaches the lowest level (PPL: 1.89). Crucially, the model exhibits higher confidence when refusing truly unknown facts ($-0.63$) compared to erroneously refusing known facts ($-0.76$). This probabilistic gap proves that the model has internalized its knowledge boundaries, enabling it to distinguish between justified abstention and unnecessary refusal at a foundational level.

\textbf{(2) Quantifying the Alignment Tax via Probabilistic Pressure.} 
In the ``Known'' (IK) category, the Logprob of the refusal template also rises from $-0.88$ to $-0.76$. This rise provides probabilistic evidence of the ``alignment tax'': as the model is incentivized to be truthful, its internal probability mass for refusal increases even for mastered facts. However, the refusal PPL for IK remains higher than that for IDK (2.14 vs. 1.89), suggesting the model maintains a healthy internal skepticism toward abstaining from questions it actually knows. This nuanced calibration explains why preference-aware optimization preserves higher IK-IK rates compared to the blanket refusal prior typically learned by SFT.

\begin{table}[t]
\centering
\caption{\textbf{Internal Uncertainty Analysis on Visual-Idk.} We analyze the average log-probability (Logprob) of generated answers and the Perplexity (PPL) of the refusal template. Pre-Align shows the intrinsic state of the model, while Post-Align reflects the ORPO aligned behavior.}
\label{tab:uncertainty_analysis}
\begin{tabular}{l|cc|cc}
\toprule
\multirow{2.5}{*}{\textbf{Category}} & \multicolumn{2}{c|}{\textbf{Answer Logprob ($\uparrow$)}} & \multicolumn{2}{c}{\textbf{Refusal PPL ($\downarrow$)}} \\
\cmidrule(lr){2-3} \cmidrule(lr){4-5}
& Pre-Align & Post-Align & Pre-Align & Post-Align \\
\midrule
\textbf{Known Questions} & -0.88 & -0.76 & 2.42 & 2.14 \\
\textbf{Unknown Questions} & -0.80 & \textbf{-0.63} & 2.22 & \textbf{1.89} \\
\bottomrule
\end{tabular}
\vspace{-3mm}
\end{table}

In conclusion, these results provide robust evidence that our Visual-Idk alignment genuinely recalibrates the model's epistemic uncertainty. The assistant does not just learn to repeat a template; it learns to allocate its probability mass strategically based on its internal knowledge mastery.

\subsection{Case Study}

\begin{figure}[t]
    \centering
    \includegraphics[width=1.0\linewidth]{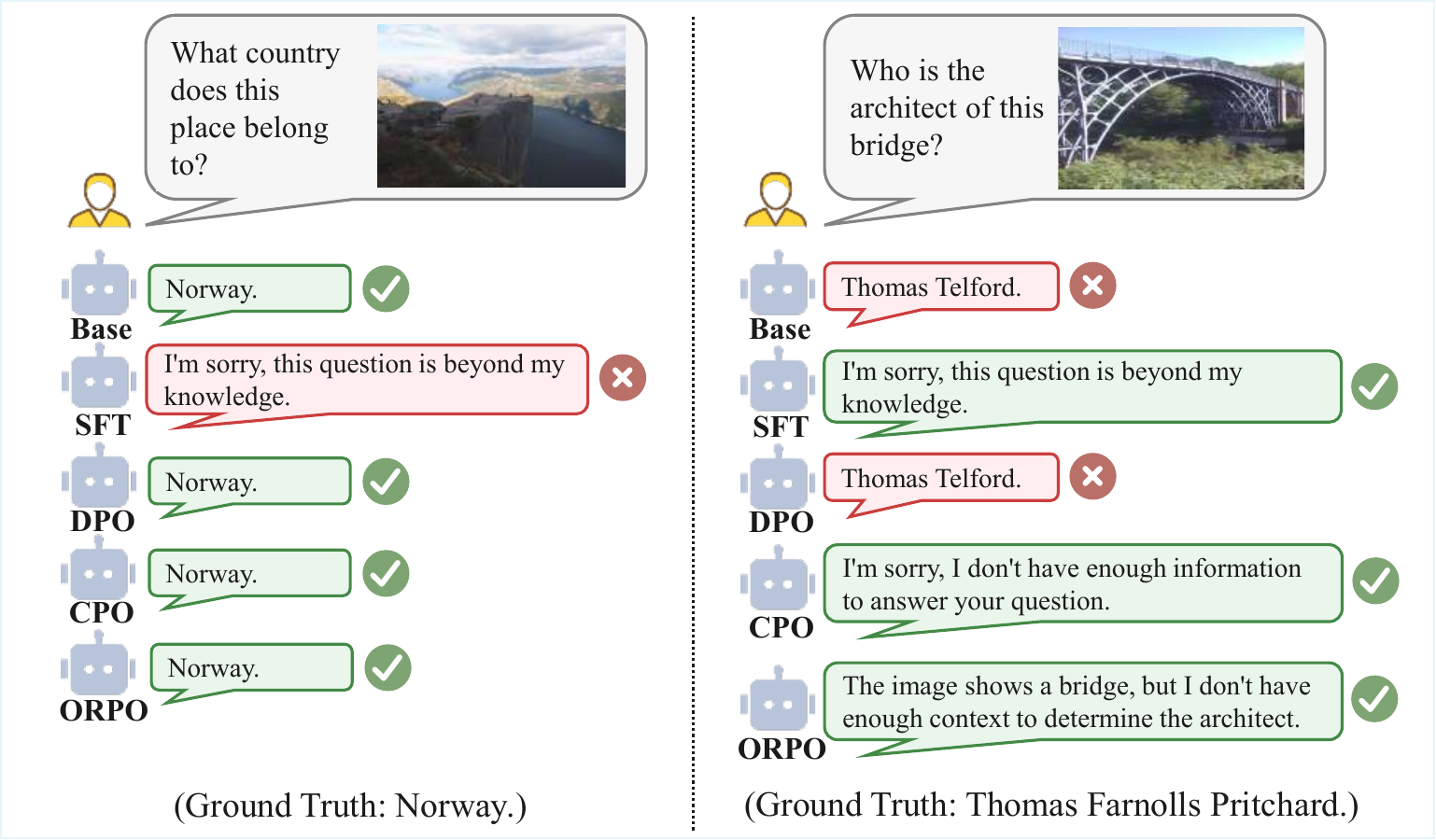}
    \caption{\textbf{ Examples of model responses before and after alignment.} The examples illustrate model behaviors on a mastered fact (\textbf{Left}) and an unmastered fact (\textbf{Right}) from the Visual-Idk test set, comparing the base LLaVA-1.5 with various alignment methods.}
    \label{fig:case_study}
\end{figure}

To provide a more intuitive understanding of how different alignment strategies reshape the model's knowledge boundaries, we present qualitative examples from the Visual-Idk test set in Fig.~\ref{fig:case_study}. These cases highlight the transition from blind overconfidence to prudent abstention. We present qualitative examples in Fig.~\ref{fig:case_study} to illustrate the behavioral shifts after alignment.

\textbf{Mitigating the Alignment Tax.} 
As shown in the left example, the base model possesses the knowledge of the location (Norway). While SFT induces excessive conservatism and erroneously refuses to answer, preference optimization methods successfully maintain helpfulness by preserving this mastered knowledge. This validates our framework's ability to refine the decision boundary without sacrificing existing utility.

\textbf{Suppressing Hallucinations.} 
In the right example, the model encounters a specialized entity beyond its mastery. The Base and DPO models generate confident but incorrect responses (``Thomas Telford''). In contrast, CPO and ORPO demonstrate superior epistemic awareness by recognizing the knowledge gap and electing to abstain. Notably, ORPO provides a more context-grounded refusal, suggesting a more calibrated internal understanding of its knowledge limits.

These cases confirm that V-Idk alignment effectively transforms VLMs from overconfident guessers into prudent visual assistants, promoting the development of more reliable and trustworthy multimodal AI systems.

\section{Conclusion}
\label{sec:conclusion}
This paper presents the Visual-Idk alignment framework to delineate the knowledge boundaries of VLMs. Specifically, we curate a model-specific preference dataset through a systematic three-stage construction pipeline, followed by alignment training using diverse strategies. We conduct extensive Experiments on multiple datasets. The results on the Visual-Idk dataset demonstrate that this framework effectively transforms factual hallucinations into honest abstentions while preserving mastered knowledge and mitigating the ``alignment tax.'' This learned honesty generalizes to out-of-distribution visual and knowledge shifts datasets, as further validated by internal probabilistic probing. Overall, our work offers a robust methodology for developing prudent visual assistants, contributing to the broader pursuit of more reliable and trustworthy multimodal systems. As for limitations, the number of training data is kind of small, and further work could enrich the training data for better exploration.

\bibliographystyle{splncs04}
\bibliography{main}
\end{document}